%% file: main.tex
\documentclass[jair,twoside,11pt,theapa]{article}
\usepackage[implicit=false]{hyperref}
\usepackage{jair, theapa, rawfonts}
\usepackage{url}
\usepackage{subcaption}
\usepackage{todonotes}

\usepackage[utf8]{inputenc} 
\usepackage[T1]{fontenc}    
\usepackage{amsfonts}       
\usepackage{nicefrac}       
\usepackage{microtype}      
\usepackage{xcolor}         
\usepackage{booktabs}       
\usepackage{graphicx}
\usepackage{amsmath,bm}
\usepackage{amsthm, amssymb, dsfont}
\usepackage{cleveref}
\usepackage{multirow}
\usepackage{makecell}
\usepackage{pgfplots}
\DeclareUnicodeCharacter{2212}{−}
\usepgfplotslibrary{groupplots,dateplot}
\usetikzlibrary{patterns,shapes.arrows}
\pgfplotsset{compat=newest}

\jairheading{79}{2024}{639-677}{03/2023}{02/2024}
\ShortHeadings{Guidelines and Opportunities for Fairness-Aware AutoML}
{Weerts, Pfisterer, Feurer, Eggensperger et al.}
\firstpageno{639}

\input{macros}
\theoremstyle{definition}

\usepackage{tikz}
\usetikzlibrary{arrows.meta, decorations.pathreplacing, calligraphy, shadows.blur, patterns, patterns.meta, shapes.geometric}
\tikzset{narrowtriangle/.tip={Triangle[length = 2pt 1.5, width=4pt 2, round, line width = 1pt 1]}}

\pgfplotsset{every axis/.append style={
      axis line style={ultra thick, -Latex}, 
      xlabel style={font=\LARGE\sffamily},
      ylabel style={font=\LARGE\sffamily},
      label style={font=\LARGE},
      tick label style={font=\LARGE},
      legend style={font=\LARGE},
    }
}
\begin{document}

\title{Can Fairness be Automated? Guidelines and Opportunities for Fairness-aware AutoML} 
\author{%
\name Hilde Weerts
\email h.j.p.weerts@tue.nl \\
\addr Eindhoven University of Technology
\AND
\name Florian Pfisterer
\email pfistererf@googlemail.com \\
\addr Ludwig-Maximilians-Universität München \\ Munich Center for Machine Learning
\AND
\name Matthias Feurer 
\email feurerm@cs.uni-freiburg.de \\
\addr Albert-Ludwigs-Universität Freiburg
\AND
\name Katharina Eggensperger 
\email katharina.eggensperger@uni-tuebingen.de \\
\addr Albert-Ludwigs-Universität Freiburg \\ University of Tübingen
\AND
\name Edward Bergman 
\email bergmane@cs.uni-freiburg.de \\
\name Noor Awad 
\email awad@cs.uni-freiburg.de \\
\addr Albert-Ludwigs-Universität Freiburg 
\AND
\name Joaquin Vanschoren
\email j.vanschoren@tue.nl \\
\name Mykola Pechenizkiy
\email m.pechenizkiy@tue.nl \\
\addr Eindhoven University of Technology
\AND
\name Bernd Bischl 
\email bernd.bischl@stat.uni-muenchen.de \\
\addr Ludwig-Maximilians-Universität München\\Munich Center for Machine Learning
\AND
\name Frank Hutter 
\email fh@cs.uni-freiburg.de \\
\addr Albert-Ludwigs-Universität Freiburg
}

\maketitle

\begin{abstract}
The field of automated machine learning (AutoML) introduces techniques that automate parts of the development of machine learning (ML) systems, accelerating the process and reducing barriers for novices.
However, decisions derived from ML models can reproduce, amplify, or even introduce unfairness in our societies, causing harm to (groups of) individuals. 
In response, researchers have started to propose \automl{} systems that jointly optimize fairness and predictive performance to mitigate fairness-related harm.
However, fairness is a complex and inherently interdisciplinary subject, and solely posing it as an optimization problem can have adverse side effects.
With this work, we aim to raise awareness 
among developers of \automl{} systems 
about such limitations 
of \fautoml{},
while also calling attention to the potential of \automl{} as a tool for fairness research.
We present a comprehensive overview of different ways in which fairness-related harm can arise and the ensuing implications for the design of \fautoml{}. We conclude that while fairness cannot be automated, \fautoml{} can play an important role in the toolbox of ML practitioners. 
We highlight several open technical challenges for future work in this direction. Additionally, we advocate for the creation of more user-centered assistive systems designed to tackle challenges encountered in fairness work.

\end{abstract}

\section{Introduction}\label{sec:intro}
Machine learning (ML) is a game-changing technology that has disrupted modern data-driven applications and is increasingly deployed in various applications and contexts. However, ML systems may reproduce, amplify, or even introduce unfairness in our society, causing harm to (groups of) individuals. Examples range from facial recognition systems that disproportionately fail for darker-skinned women~\shortcite{buolamwini-facct18a}, gender bias in automatic captions~\shortcite{tatman-eacl2017a} and resume parsing~\shortcite{amazon-recruiting}, to the underestimation of the healthcare needs of black patients~\shortcite{obermeyer-science19a}. In response to the growing need for ML systems that align with principles of fairness, researchers have proposed numerous techniques to assess and mitigate unfairness of ML systems. Additionally, several open-source software libraries facilitate the application of these methods~\shortcite{aif360,fairlearn}. In practice, however, it can be challenging to incorporate fairness considerations in the design and development of ML systems, due to the following obstacles:
\begin{enumerate}
    \item \textit{Potential fairness-related harms are rarely prioritized} due to a lack of awareness, difficulties in anticipating fairness-related harm, or difficulties in advocating for required resources within an organization~\shortcite{madaio-cscw22a}.
    \item \textit{Practitioners are overwhelmed with a multitude of fairness metrics and interventions}. Choosing an appropriate metric or intervention is an active area of research and is further complicated by practical obstacles, such as the feasibility of collecting more (representative) data, the accessibility of descriptive meta-data, and the availability of cultural, contextual, and domain knowledge~\shortcite{holstein-chi19a}.
    \item \textit{Once found, solutions rarely carry over to new problems}. Given the complexity and diversity of systems, contexts, and use and failure cases, fairness-aware models cannot be repurposed within a different context~\shortcite{selbst-fat19a}.
\end{enumerate}
\paragraph{The case for \fautoml{}}
The rising field of \automl{}~\shortcite{hutter-book19a} focuses on reducing the complexity inherent to applying ML algorithms in practice by providing methods that partially automate the ML workflow. In response to fairness concerns, recent research has started to propose \automl{} systems that take into account fairness objectives.
By lowering the barrier to incorporate fairness considerations in an ML workflow, \automl{} systems might be able to partially address several of the aforementioned problems, as they reduce the ML expertise required to build, interrogate and evaluate fairness-aware ML systems:
\begin{enumerate}
    \item \emph{\Fautoml{} can reduce the barrier to entry.} It can substantially reduce turnaround times for developing and evaluating models along different metrics, lowering the amount of required resources, time, and technical expertise required to engage in fairness work, and enabling an iterative preference elicitation process for determining what precisely it means to be fair in a particular application.
    \item \emph{\Fautoml{} can facilitate a declarative interface} for domain experts to specify \emph{what} they would like to achieve, rather than \emph{how} to do so. This relieves them of the necessity of staying up to date with the latest technical fairness interventions.
    \item \emph{In contrast to individual fairness-aware models, the \fautoml{} process for finding such models carries over to new problems much better.} This is the entire point of \automl{}: finding custom technical solutions for new problems that best satisfy the problem's particular desiderata. 
\end{enumerate}
However, fairness is an intricate and inherently interdisciplinary subject. Existing fairness-aware ML algorithms often operate under the assumption that fair outcomes can be achieved by optimizing for an additional fairness metric, but reality is much more complex. In particular, the scope of these algorithms is typically limited, while potential harm can originate at each step of the ML workflow. Considering fairness-aware \textit{Auto}ML, we have a potentially more flexible and powerful tool to aid practitioners in fairness work. However, care should be taken to avoid replicating or even exacerbating known issues with existing approaches in fairness-aware ML.
This raises the question: \textit{can fairness be automated?}

\paragraph{Contributions}
With this work, we aim to raise awareness on the opportunities and limitations of \emph{\fautoml{}}.\footnote{To highlight that \automl{} is not a panacea for algorithmic fairness, and to emphasize that employing \textit{fair AutoML} systems does not automatically ensure fair solutions, we recommend referring to \automl{} systems that incorporate fairness considerations as \emph{fairness-aware} \automl{}, rather than \emph{fair} \automl{}.} Our main contribution is a comprehensive overview of important challenges in fairness-aware ML as they apply to \automl{} and the ensuing implications for the design of \fautoml{} systems. Additionally, we highlight several challenges and opportunities unique to fairness-aware \emph{Auto}ML and lay a foundation for future work toward more user-centered assistive systems. We focus on supervised learning tasks, noting that generative AI shares many of the challenges we outline but requires additional considerations.
With this work, we hope to start a discussion among the research communities of fairness and \automl{} to jointly work towards mitigating bias and harm in ML.

\paragraph{Outline}
The remainder of this article is structured as follows. Section~\ref{sec:background} covers the relevant background knowledge on algorithmic fairness and \automl{} and provides a review of existing work in the domain of fairness-aware AutoML. We then cover existing challenges and their implications along two key perspectives: user input and interaction (\Cref{sec:user}) and the design of the \automl{} system itself (\Cref{sec:automl_config}). \Cref{sec:opportunities} then identifies opportunities that arise from the use of AutoML in fairness contexts. ~\Cref{sec:conclusion} concludes with a set of guidelines for the design \& use of \fautoml{} and an inventory of directions for future work.

\section{Background}
\label{sec:background}
To better understand the role of \automl{} in a fairness-aware ML workflow, we first need to determine what such a workflow looks like. Figure~\ref{fig:pipeline} presents a typical ML workflow consisting of five highly iterative stages:
\begin{enumerate}
    \item \textit{Problem understanding.}\footnote{In the original CRISP-DM process model, \shortciteA{shearer-jdw00a} refer to this stage as \textit{business} understanding. To emphasize that ethical concerns are central to responsible design, rather than an add-on to business imperatives, we refer to the stage as \textit{problem} understanding.} The problem is scoped and translated into an ML task including real-world success criteria, requirements, and constraints.
    \item \textit{Data understanding and preparation.}\footnote{In the original CRISP-DM model, data understanding and data preparation are separate stages. For ease of presentation, we have consolidated these in one step.} It is determined which data is required, followed by an iterative sequence of data collection, exploration, cleaning, and aggregation.
    \item \textit{Modeling}. The model selection pipeline is designed, and candidate models are evaluated against technical performance metrics and constraints, with a focus on factors such as accuracy and generalization.
    \item \textit{Evaluation}. The selected model is evaluated more broadly against real-world success criteria and requirements, including, e.g., A/B tests, user tests, or adversarial tests. Note that this evaluation goes beyond the technical evaluation performed during the \textit{modeling} stage and explicitly considers whether the model meets real-world objectives.
    \item \textit{Deployment}. The deployment, monitoring, and maintenance are planned and executed. 
\end{enumerate}
In the remainder of this section, we provide an introduction to algorithmic fairness and \automl{} and highlight how each of them plays a part in the ML workflow.
We focus on supervised learning tasks with access to group memberships.
\input{figures/workflownew.tex}

\subsection{Algorithmic Fairness} 
The goal of algorithmic fairness is to ensure that the real-world outcomes of ML systems are fair with respect to legal, regulatory, or ethical notions of fairness. This is particularly relevant in the context of automated decision-making, where ML models are used to assist decision-makers or even obtain decisions automatically.

\subsubsection{Fairness Metrics}
Algorithmic fairness metrics typically measure the extent to which some form of equality is violated. The metrics differ primarily in terms of \textit{what} should be equal. In the context of supervised learning, the two most prominent notions of fairness are group fairness and individual fairness. \textit{Group fairness} is a notion of fairness that requires particular group statistics to be equal across (sub)groups defined by \textit{sensitive features}. Sensitive features intend to measure characteristics of individuals for which disparate outcomes based on that characteristic are considered undesirable from an ethical or legal point of view. Typical examples are age, sex, disability status, ethnic origin, race, or sexual orientation. Researchers have defined various group fairness metrics that differ in terms of which group statistic should be equal, typically involving either the distribution of predicted outcomes (e.g. selection rate, average predicted value) or the predictive performance of the model (e.g. accuracy, precision, recall).

Metrics of \textit{individual fairness} take the perspective of the individual, inspired by the Aristotelian principle that ``like cases should be treated alike''. Individual fairness metrics differ primarily in terms of what is considered `similar'. Statistical interpretations quantify the similarity of two instances in terms of a similarity metric~\shortcite{dwork-itcsc12a} that aims to capture similarity on task-relevant characteristics (which typically excludes sensitive characteristics). Contrarily, counterfactual fairness~\shortcite{kusner-neurips17a} takes a causal perspective, requiring similar treatment for an individual and the counterfactual of that same individual, had they belonged to a different sensitive group.

Most research has focused on group fairness. In particular, the two most prominent group fairness metrics are \textit{demographic parity} and \textit{equalized odds}. We denote with random variables the predicted outcome $R$, the true outcome $Y$, and the membership of a sensitive group $A$.

\paragraph{Demographic Parity} 
Demographic parity~\shortcite{calders-dm10a} requires that the probability of predicting the positive class is independent of the sensitive class $A$: $P(R=1|A=a)=P(R=1)$, essentially requiring independence~\shortcite{barocas-fairmlbook} between the sensitive feature and the predicted outcome: $R \perp A$. Demographic parity is satisfied when the selection rate is equal across groups. For example, in a resume selection scenario, demographic parity holds if the proportion of selected resumes is the same for each sensitive group. 

Demographic parity does not take into account the true label $Y$. If base rates are different across groups, i.e., $P(Y=1|A=a) \neq P(Y=1)$, satisfying demographic parity requires one to make predictions that do not coincide with the observed outcomes, meaning that demographic parity rules out a perfect predictor. The metric can be classified as what \shortciteA{wachter-vlr2020a} refers to as a bias-\textit{transforming} metric: optimizing for this metric corresponds to changing the (observed) status quo. 

\paragraph{Equalized Odds}
Equalized odds~\shortcite{hardt-nips16a} is one of the most commonly studied fairness metrics that requires an equal distribution of errors across sensitive groups. In particular, it asks for equal true and false positive rates across groups: $P(R=1|Y=1,A=a)=P(R=1|Y=1) \land P(R=1|Y=0,A=a)=P(R=1|Y=0)$. This essentially requires independence between the sensitive feature and the predicted outcome, conditional on the true outcome: $R \perp A \,|\, Y$~\shortcite{barocas-fairmlbook}.

As opposed to demographic parity, equalized odds does explicitly take into account $Y$. It is therefore what \shortciteA{wachter-vlr2020a} refers to as a bias-\textit{preserving} metric: optimizing for equalized odds will preserve the status quo as much as possible, implicitly assuming that any bias present in the data should be preserved.

\subsubsection{Fairness-Aware Machine Learning Algorithms}
Interventions to mitigate fairness-related harm can take place during all stages of an ML workflow~\shortcite{mitchell-ars21a,mehrabi-acm21a,raji-facct20a,madaio-chi20a}. Such interventions include ethical reviews of a problem definition, identifying and mitigating potential biases in the data, exhaustive fairness assessments during evaluation and deployment, and active stakeholder involvement during all stages (see Figure~\ref{fig:pipeline}).
However, the vast majority of the algorithmic fairness literature has focused on fairness-aware ML algorithms applied during the modeling stage ~\shortcite{holstein-chi19a}. Such technical interventions formulate fairness as an optimization task, where the goal is to achieve high predictive performance whilst also satisfying a fairness constraint.
They can be roughly subdivided into three categories~\shortcite{kamiran-discrimination13a}.

\paragraph{Pre-processing} Pre-processing approaches can adjust the data to obscure any undesirable associations between sensitive features and a target variable~\shortcite<e.g.,>{kamiran-kis11a}. 

\paragraph{Constrained learning} Constrained learning techniques directly incorporate a fairness constraint in the learning algorithm, either by adapting existing learning paradigms~\shortcite<e.g.,>{calders-icdm09a,zafar-aistat17a} or through wrapper methods~\shortcite<e.g.,>{agarwal-icml18a}.

\paragraph{Post-processing} Post-processing approaches adjust a trained ML model, either through post-processing predictions~\shortcite<e.g.,>{hardt-nips16a} or by adjusting the model parameters directly~\shortcite<e.g.,>{kamiran-icdm10a}.\\

\noindent We refer to~\shortciteA{caton-acm23a} for a more elaborate overview of existing approaches. There is currently little guidance on how to select fairness-aware ML techniques, their effectiveness across scenarios and in what cases their use is appropriate. 

\subsection{Automated Machine Learning}
\automl{}~\shortcite{hutter-book19a,escalante-ncs21a} is a subfield of machine learning that researches and studies methods that automate components of the ML workflow with the goals of speeding up the development of ML applications and reducing the required level of expertise for otherwise manual tasks. By lowering the barrier to apply ML methods, ML can become more accessible and allows to explore new use cases.  \automl{} comes in many flavours which we briefly describe in the following. 

\paragraph{Hyperparameter optimization (HPO)}
The most low-level incarnation of \automl{} is HPO, where the goal is to optimize the hyperparameters of an ML algorithm to minimize a user-specified cost function, such as the misclassification error~\shortcite{feurer-automlbook19a}. HPO is often used to tune the hyperparameters of deep neural networks~\shortcite{bergstra-nips11a,snoek-nips12a,snoek-icml15a} and popular techniques are grid and random search~\shortcite{bergstra-jmlr12a,bouthillier-hal20a}, Bayesian optimization~\shortcite{garnett-bayesoptbook22a} and bandit-based methods~\shortcite{jamieson-aistats16a,li-jmlr18a}.

The \emph{Combined Algorithm Selection and Hyperparameter optimization problem} (CASH problem) extends the HPO problem to optimize an ML pipeline choosing between different ML methods~\shortcite{thornton-kdd13a}:
\begin{equation}
    A^*, \conf^* \in \argmin_{A^{(j)} \in \mathcal{A},\conf \in \bm{\Lambda}^{(j)}} \, \frac{1}{k}\, \sum^{k}_{i=1} \, \mathcal{L}(A_{\conf}^{(j)}, \mathcal{D}^{(i)}_{train}, \mathcal{D}^{(i)}_{test}).
\label{eq:cash}
\end{equation}
It is a hierarchical hyperparameter optimization problem with the hyperparameter space $\bm{\Lambda} = \bm{\Lambda}^{(1)} \cup \dots \cup \bm{\Lambda}^{(l)} \cup \Lambda_r$, where $\lambda_r \in \Lambda_r = \{A^{(1)}, \dots, A^{(l)}\}$ is a new root-level hyperparameter that selects between algorithms $A^{(1)}, \dots, A^{(l)}$. The \textit{search space} consists of subspaces $\bm{\Lambda}^{(j)}$ that are conditional on $\lambda_r$ being instantiated to $A_j$. This setting is often extended to contain multiple high-level choices $r \in (1, \dots, R)$ to construct a multi-step pipeline.
The CASH problem requires specialized HPO algorithms that can deal with these hierarchical hyperparameters, for example, tree-powered Bayesian optimization methods~\shortcite{hutter-lion11a,bergstra-nips11a,rakotoarison-ijcai19a}, or decompositions into higher and lower-level problems~\shortcite{liu-aaai20a}.

\paragraph{Neural architecture search (NAS)}
The recent success of deep learning methods has spurred the development of tailored methods that optimize the structure of deep neural networks~\shortcite{elsken-automlbook19a,white-arxiv23a}. Closely related to HPO, some NAS problems can be tackled with HPO methods. Two popular approaches to NAS are gradient-based optimization via continuous relaxations of the architectures~\shortcite{liu-iclr19a} and the global optimization of repeated local structures in a so-called cell search space~\shortcite{real-aaai19a}.
The NAS and HPO problems can be combined into the joint
architecture and hyperparameter search (JAHS)
problem, in which both the topology of the neural network and its hyperparameters such as the learning rate and regularization hyperparameters, are optimized together~\shortcite{zela-automl18a,zimmer-ieee21a,hirose-acml21a,bansal-neurips22a}.

\paragraph{AutoML systems} 
\automl{} systems typically extend the above-mentioned techniques with a concrete design space to automatically design ML pipelines. In its basic form, an \automl{} system for supervised learning takes as input a \textit{dataset} and an \textit{evaluation metric} and outputs an ML pipeline optimized for the metric. As such, the \textit{scope} of \automl{} is mostly limited to the \textit{modeling} stage of a machine learning development workflow (Figure~\ref{fig:pipeline}). 

\automl{} systems primarily differ in terms of the considered \textit{search space} and the \textit{search strategy} that is used to explore the search space. Popular examples are AutoWEKA~\shortcite{thornton-kdd13a} that uses the Bayesian optimization algorithm SMAC~\shortcite{hutter-lion11a} to solve the CASH problem given in Equation~\ref{eq:cash}, the extension Auto-sklearn~\shortcite{feurer-nips15a} that also incorporates meta-learning and explicit ensembling, and the earlier Particle Swarm Model Selection~\shortcite{escalante-jmlr09a} that uses a particle swarm optimization algorithm.
In contrast, the TPOT \automl{} system~\shortcite{olson-gecco16a} uses genetic programming and can construct more general pipelines. These \automl{} systems are composed of a general search space description of an ML pipeline and its hyperparameters. They then employ powerful global HPO algorithms~\shortcite{feurer-automlbook19a} to optimize the pipeline for a dataset at hand, often using iterative optimization algorithms that suggest and refine solutions during the optimization process. However, there are also other approaches that do not rely on HPO, such as Auto-Gluon~\shortcite{erickson-arxiv20a} which stacks models to yield a powerful predictive pipeline. We refer to \shortciteA{gijsbers-arxiv22a} and \shortciteA{feurer-jmlr22a} for recent overviews of \automl{} systems. In addition to these, the broader field of \emph{automated Data Science} (AutoDS;~\shortciteA{deBie-cacm22a}) focuses on automating additional aspects of the development process, such as exploratory data analysis, data preparation, or parts of model deployment. The focus of the current paper is on \automl{}, but many of our findings are applicable to other endeavors related to automation in machine learning, including AutoDS.

\subsection{AutoML and Fairness}
\label{sec:fautoml}

As the interest in fairness-aware ML systems increases, researchers have started to propose AutoML systems that incorporate fairness considerations. In this section we discuss multi-objective and constrained optimization, then describe how it is used to handle fairness as an additional objective, and lastly, highlight other findings that relate to fairness and \automl{}.

\paragraph{Multi-objective and constrained optimization}
HPO methods, \automl{} systems, and NAS methods have been extended to the \textit{multi-objective optimization} case to take into account additional objectives.
A related solution is given by \textit{constrained optimization}, in which one or more objectives are optimized to meet a certain constraint value~\shortcite{lobato-jmlr16a}.

HPO has been used to optimize ML models to take multiple cost functions~\shortcite{horn-ssci16a} or resource consumption~\shortcite{igel-emco05a} into account. The goal of multi-objective optimization is to approximate the Pareto set representing the best possible combination of outcomes -- solutions that can not be improved \wrt{} one objective without sacrificing performance \wrt{} another. The \automl{} system TPOT uses multi-objective optimization to balance pipeline size and performance~\shortcite{olson-gecco16a}.
AutoXGBoostMC proposes tuning several additional metrics, such as fairness, robustness, or interpretability metrics~\shortcite{pfisterer-arxiv19a}. The NAS technique LEMONADE~\shortcite{elsken-iclr19a} searches for neural architectures that balance performance, latency and size. We refer to~\shortciteA{moraleshernandez-air22a} and~\shortciteA{karl-acm23a} for reviews of multi-objective HPO and to~\shortciteA{benmeziane-arxiv21a} for a review on hardware-aware NAS, a subfield of multi-objective NAS.

\paragraph{Fairness as a constraint or additional objective}
Many fairness-aware ML algorithms consider a constrained optimization approach.
Existing work on \fautoml{} has extended this approach to \automl{} systems. For example,~\shortciteA{liu-aaai20a} propose a new CASH algorithm that decomposes the CASH problem into multiple simpler subproblems and allows incorporating black-box constraints along with the optimization objective. Similarly, in fairBO, ~\shortciteA{perrone-aies21a} proposed to use standard constrained Bayesian optimization to optimize for predictive performance subject to fairness constraints, and demonstrated how Bayesian optimization can tune hyperparameters constrained for three different metrics at the same time.

Early examples of multi-objective HPO for fairness~\shortcite{pfisterer-arxiv19a,chakraborty-ase19a} use the Bayesian optimization algorithm ParEGO~\shortcite{knowls-evoco06a} and a custom sequential model-based optimization algorithm~\shortcite{nair-ieee20a}, respectively. Later works propose to use multi-fidelity optimization and extend the popular Successive Halving~\shortcite{karnin-icml13a,jamieson-aistats16a} and Hyperband~\shortcite{li-jmlr18a} algorithms to multi-objective algorithms and use them to tune fairness as an auxiliary metric~\shortcite{schmucker-metalearn20a,schmucker-arxiv21a,cruz-icdm21a}. Most recently, \shortciteA{dooley-neurips23a} used multi-fidelity multi-objective joint neural architecture and hyperparameter optimization in order to find models that perform better with respect to various fairness metrics in face recognition.

Finally, \emph{Fair AutoML}~\shortcite{wu-arxiv21a} extended the FLAML \automl{} system~\shortcite{wang-mlsys21a} by adding a resource allocation strategy to dynamically decide whether to evaluate new pipelines to improve performance or to mitigate bias for promising pipelines by using the exponentiated gradient technique~\shortcite{agarwal-icml18a}.

\paragraph{Gaining knowledge through \fautoml{}}
Additionally, \shortciteA{perrone-aies21a} thoroughly study the impact of tuned hyperparameters on a fairness metric and find that larger regularization often leads to fairer results. Moreover, they find that tuning standard ML algorithms leads to comparable or even better performance than tuning fairness-aware ML algorithms, such as support vector machines that incorporate a constraint on a fairness metric during model training~\shortcite{donini-neurips18a}. 
Similarly, \shortciteA{dooley-neurips23a} demonstrate that by only optimizing the neural architecture and hyperparameters of deep neural networks, it is possible to yield better combinations of fairness metric and predictive performance than by relying on existing bias mitigation approaches, and that fairer neural architectures exhibit a reduced linear separability of protected attributes.
Finally, \shortciteA{cruz-icdm21a} find that targeted hyperparameter optimization can improve fairness metrics at a small cost in predictive performance. Moreover, they also find that different ML algorithms occupy distinct regions of the fairness-predictive performance space, suggesting a relationship between specific model classes and fairness.\\

\noindent In conclusion, existing work on \fautoml{} has combined fairness-aware ML techniques proposed in algorithmic fairness research, as well as the choice of neural architectures and hyperparameters, with multi-objective or constrained optimization. This offers several potential advantages over existing techniques, e.g. in terms of flexibility. However, as the research field of algorithmic fairness has evolved to include more (inter)disciplinary perspectives, it has become clear that formulating fairness as an optimization task is not always effective in achieving fairer real-world outcomes.
This raises the question: how should we design \fautoml{} systems to achieve fair outcomes? In the remainder of this work, we will dive deeper into the limitations and opportunities of \fautoml{} from two perspectives: the user's inputs into the system
(Section~\ref{sec:user}) and the \automl{} system's design (Section~\ref{sec:automl_config}).

\section{Effect of the User Inputs and User Interactions on Fairness}
\label{sec:user}
\automl{} - and by extension \fautoml{} - traditionally mostly addresses the \textit{modeling} stage of an ML workflow. However, fairness-related harm can arise at each stage of the ML development process. Therefore, the way in which an \automl{} system is integrated into a workflow plays an important part in achieving fair real-world outcomes. In this section, we set out how the user's input (data, metrics, and constraints)
affects fairness and what this implies for \fautoml{}.

\subsection{Data}
\label{sec:data}
Data is one of the main user inputs to an \automl{} system. Coincidentally, biases in datasets are typically regarded as one of the primary sources for fairness-related harm, which makes bias identification and mitigation crucial components of a fairness-aware ML workflow (Figure~\ref{fig:pipeline}). There are many different types of bias that can be present in datasets. We will limit our overview to three biases that are particularly important for notions of group fairness and refer the interested reader to~\shortciteA{mitchell-ars21a} and~\shortciteA{mehrabi-acm21a} for a more exhaustive overview.

\paragraph{Historical Bias}
\textit{Historical bias} refers to social biases that are encoded in a dataset and can be reproduced by an ML model~\shortcite{suresh-eaamo21a}. In particular, a dataset may reflect historical injustices that exist in our societies~\shortcite{bao-neuripsdbt21}. For example, if historically people from lower socioeconomic backgrounds have had fewer opportunities to receive a high-quality education, they may be less suitable for jobs where education is essential, resulting in lower hiring rates for these groups. Unaccounted for, an ML model is likely to reproduce this type of bias and predict fewer positives for historically marginalized groups.

\paragraph{Measurement Bias}
Datasets embed a plethora of subjective design choices, including what is measured and how. \textit{Measurement bias} occurs when measurements do not accurately or adequately measure the concept we intended to measure. When measurement bias is related to sensitive group membership, it can be a source of unfairness. For example, recidivism risk assessment models are sometimes trained on arrest records, which reflect only a subset of all criminal behavior and might be affected by biased policing practices. As a result, arrest records can be a biased measurement of true criminal activity.  
Importantly, these issues cannot be observed from the data alone and must be inferred from the context. As a result, meaningfully addressing measurement bias will require careful consideration of the data generation and collection processes~\shortcite{jacobs-facct21a}.

\paragraph{Representation Bias} 
A dataset may suffer from \emph{representation bias},\footnote{Despite apparent similarities, representation bias is not the same as selection bias in statistical analysis. Selection bias refers to a failure to achieve proper randomization in selection samples, while representation bias refers to the minimum number of samples that are required to make accurate predictions for a subgroup.}
meaning that the dataset contains too few instances of a sensitive group to make accurate predictions. This issue is most prevalent when the data distribution differs substantially between groups. For example, \shortciteA{buolamwini-facct18a} have shown that facial recognition systems failed disproportionately for pictures of darker-skinned women, a group consistently underrepresented in facial recognition datasets. Even if the number of instances is similar, features included in a dataset may be less informative for some groups compared to others~\shortcite{barocas-clr16a}.

\paragraph{Implications}
While it is impossible to identify all types of bias from data alone, future \fautoml{} systems can include several safeguards and diagnostic tools that support practitioners. In particular, systems can flag potential issues, such as disparate base rates (pointing to historical or measurement bias) and small (sub)groups (pointing to representation bias). This can then be used to trigger a user interaction. Furthermore, future work could explore incorporating approaches that facilitate careful modeling of historical bias and measurement bias through causal or statistical modeling, assisted by \automl{} methods. 
Additionally, future systems could incorporate requests for meta-data in the user interface~\shortcite<e.g., in the form of datasheets;>{gebru-acm21a} that help practitioners identify potential issues with their data and facilitate documentation for audits.

\subsection{Selecting Fairness Metrics}\label{sec:selecting_fairness}
Another important user input of an \automl{} system is the metric(s) for which the ML model is optimized.
As seen in Section~\ref{sec:fautoml}, \fautoml{} systems typically use a multi-objective or constrained learning approach to incorporate quantitative fairness metrics in the learning process. An underlying assumption of this approach is that all relevant fairness and performance metrics can be clearly identified and adequately reflect all relevant ethical considerations (ethical review and system (re-)design in Figure~\ref{fig:pipeline}). This assumption is unlikely to hold in practice for several reasons.

\paragraph{Identifying relevant groups is challenging} 
Addressing fairness requires identifying socially relevant groups that are at risk of harm. It is often hard to anticipate in advance for which subgroups the system might fail~\shortcite{chen-facct19a,wachter-clsr21a}, or to even identify them in advance~\shortcite{ruggieri-aaai23a}, resulting in a reactive strategy rather than an anticipatory approach~\shortcite{shankar-arxiv22a}. Moreover, commonly studied sensitive characteristics, such as race and gender, are social constructs that are complex to measure~\shortcite{hanna-fat20a,jacobs-facct21a}; this is further complicated when we consider intersectionality~\shortcite{chen-facct19a,buolamwini-facct18a} and privacy concerns.

\paragraph{Choosing a fairness metric is challenging}
Determining which fairness metric is most appropriate for a given use case is a highly non-trivial question and still an active area of research~\shortcite{hellman-vlr20a,wachter-clsr21a,hertweck-facct21a,hedden-ppa21a,weerts-arxiv22a}. This is further complicated by the fact that many fairness constraints cannot be satisfied simultaneously~\shortcite{kleinberg-itcs17a,chouldechova-bigdata17a} as each fairness metric corresponds to a different set of empirical and normative assumptions~\shortcite{weerts-arxiv22a}. For example, arguments for or against demographic parity can depend both on the causes of disparities (e.g., measurement bias or historical bias) and the consequences (e.g., the cost of false positives).

\paragraph{Fairness metrics are simplified notions of fairness}
Every quantitative fairness metric is necessarily a simplification, lacking aspects of the substantive nature of fairness long debated by philosophers, legal scholars, and sociologists~\shortcite{selbst-fat19a,jacobs-facct21a,schwobel-facct22a,chen-arxiv22a}.
While simplified notions of fairness can potentially help to \emph{assess} potential fairness-related harms, it can be challenging to anticipate all relevant side effects of (technical) interventions that \emph{enforce} them. In particular, group fairness metrics are predominantly parity-based, meaning they enforce equality of statistics, such as error rates, across sensitive groups. Beyond parity, group fairness metrics do not set specific constraints on the distribution of predictions. For example, it is theoretically possible to achieve equal false positive rates by increasing the false positive rate of the better-off group such that it is equal to that of the worst-off group, which in most cases will not make the worst-off group any better off. The under-specification of fairness metrics can be especially problematic when we consider bias-\textit{transforming} metrics such as demographic parity that require deviating from the status quo. Underspecification of the world we would like to see can do more harm than good. Borrowing an example from~\shortciteA{dwork-itcsc12a}, it is possible to increase the selection rate for female applicants by inviting the least qualified female applicants for an interview. While this intervention would satisfy demographic parity, it is unlikely that any of the unqualified applicants would actually make it to the next round. Although extreme ``solutions'' like this are unlikely to end up on the Pareto front of a \fautoml{} solution, more subtle variations of this issue can be hard to detect. Without careful modeling of measurement bias and historical bias, simply enforcing demographic parity -- while a very important measure for equitable outcomes -- can have undesirable side effects that can even harm the groups the intervention was designed to protect~\shortcite{liu-icml18a,weerts-arxiv22a}.

\paragraph{Implications}
On the one hand, flexibility in defining fairness metrics is crucial to address various fairness concerns and most current \automl{} systems are indeed agnostic to the metric(s) to be optimized.\footnote{Currently, only the fairness extension of FLAML~\shortcite{wu-arxiv21a} is limited in this regard due to its use of the exponentiated gradient to obtain fairer models, which is only defined for a limited number of metrics. However, this is to a large extent a limitation of the implementation rather than the methodology.} 
On the other hand, practitioners may not always anticipate adverse side effects of such (hand-crafted) fairness metrics~\shortcite{liu-icml18a,weerts-arxiv22a} or lack available implementations, e.g. for settings other than binary classification.
This is not to say that practitioners should refrain from attempting to quantify fairness at all, which would simply lead to a continuation of current practice. 
Moreover, we also think that these are not reasons against fairness-aware \emph{Auto}ML, as all of the above problems of choosing an appropriate fairness metric similarly apply to standard ML.

Instead, practitioners need to appreciate the complexities of quantifying fairness, solicit input from relevant stakeholders, exercise caution in optimizing for fairness metrics, and proactively monitor the system for unforeseen side effects. Indeed, it is our hope that the faster development cycle that fairness-aware \emph{Auto}ML affords will facilitate more iterations on these complex issues.
While the use of limited fairness metrics will often be better than ignoring fairness entirely, developers of \fautoml{} systems need to steer clear of ``fairness-washing'' and instead design their system in a way that encourages users to thoroughly scrutinize their models. Going beyond reporting fairness metric scores, more comprehensive evaluations could include grounding metrics in real-world quantities~\shortcite{corbett-arxiv18a}, the use of disaggregated evaluations~\shortcite{barocas-aies21a} and integration of interpretability tools~\shortcite{molnar-iweeaibdmc22a}. Additionally, we observe a tension between guiding practitioners in their choice of fairness metric, whilst avoiding oversimplification of ethical concerns. While \automl{} systems are well equipped to optimize for bias-preserving notions of fairness that emphasize equal predictive performance across groups, particular care should be taken with optimization for bias-transforming metrics. 

\subsection{Use of the System}
How a \fautoml{} system and its output are used plays an important part in achieving fair real-world outcomes. 

\paragraph{Automation Bias}
One possible advantage of fairness-aware \emph{Auto}ML over more traditional fairness-aware ML techniques is that \fautoml{} allows practitioners to spend more time on other tasks, such as careful selection of training data and the right fairness metrics for the application at hand. However, as with any technology, the introduction of \fautoml{} into the toolbox of ML practitioners can have unintended consequences. In particular, there is the risk of \textit{automation bias}: the tendency to over-trust suggestions from automated systems, even in the presence of contradictory information~\shortcite{skitka-ijhcs99a}. For example, in the context of interpretability, \shortciteA{kaur-chi20a} find that ML practitioners may take the output of ML interpretability tools at face value, rather than use them as a starting point for further inquiry and understanding of the ML model. In the case of \fautoml{}, automation bias could cause practitioners to more easily adopt an ML model suggested by the system compared to hand-crafted models, without thoroughly scrutinizing the proposed solution. In other words, practitioners may expect that \fautoml{} will not just output the \textit{best} model, but also optimizes for the \textit{right} objective almost automatically. As explained above, it is challenging to identify all relevant side effects of technical interventions in advance, which makes the thorough evaluation of models crucial. This is further complicated by organizational dynamics, which may favor the rapid deployment of models \shortcite{madaio-chi20a}. Additionally, there is the risk of over-reliance on the automated system: if a \fautoml{} system is not designed to tackle an issue (e.g., intersections of multiple sensitive features), a user might be tempted to simply ignore the problem, as not doing so means they need to resort to a manual solution.\footnote{Of course, issues related to the lack of availability of particular capabilities are not limited to \automl{} systems and also apply to (fairness-aware) ML more generally.} Finally, an iterative model development stage can lead to insights into the dataset and problem at hand that could be lost as more steps are automated.

\paragraph{Role of (Fairness-Aware) (Auto)ML}
Potential fairness-related harm depends not only on the model itself but also on the sociotechnical system the model is a part of \shortcite{selbst-fat19a} and technical interventions during the modeling stage can only hope to address parts of the problem.

Sources of unfair real-world outcomes are not limited to the direct inputs and outputs of a \fautoml{} system, but also include whether and how a practitioner incorporates the system in their workflow. 
For example, the way in which a problem is formulated during the \textit{problem understanding} stage is implicitly shaped by practitioners' assumptions~\shortcite{madaio-chi20a}.
While this is not problematic by and of itself, practitioners may not be able to identify and assess their own biases~\shortcite{raji-facct20a}.
Even prioritizing some problems over others constitutes a value judgment. A municipality may use ML to detect fraud in welfare benefit applications, but could also use ML to identify citizens who are eligible for welfare but have not yet applied. Which project is pursued affects how stakeholders will be impacted. Value judgments become even more explicit during the design of requirements and key performance indicators, which dictate what the system is optimized for.
At the other end of the workflow, during \textit{deployment}, decision-makers may interact with the ML model in a way that introduces fairness-related harm. For example, in pretrial risk assessment, not the predictions of the model but the final decision of the judge determines the real-world consequences. If fairness is a requirement, it is therefore not sufficient to consider only the output of the model, but also how the predictions are used by decision-makers~\shortcite{selbst-fat19a}. Another example is the risk of reinforcing feedback loops. Most fairness assessments during the \textit{modeling} stage are performed at a single point in time. However, once deployed, decision-making systems can shape which data is collected in the future and thus can reinforce existing or even introduce bias~\shortcite{ensign-fat18a,liu-icml18a,cooper-aies21a,schwobel-facct22a}.

The introduction of fairness-aware ML, and by extension \fautoml{}, could incentivize ML practitioners to disregard other, potentially more effective, options in the decision space. 
For example, when data is taken as a given input, this negates the possibility to collect more or better data. Similarly, inquiry into the data and its provenance might lead to insights that improve the quality of a resulting model, e.g. detecting data leakage or relevant missing features.
Additionally, a non-technical intervention might be more appropriate. For example, rather than decreasing the false positive rate in recidivism risk assessments through technical interventions, false positives could be made less costly by changing the real-world consequences from sending a defendant to jail to enrolling them in a community service program~\shortcite{corbettdavies-kdd17a}.

\paragraph{Implications}
In order to avoid falling prey to the adverse effects of automation, \automl{} system developers could consider approaches introduced in the field of human-computer interaction. For example, to avoid users simply accepting the default solutions, principles of \textit{seamful design} \shortcite{kaur-facct22a} can be used to implement intentional friction that promotes critical reflection. In this way, \fautoml{} users can be encouraged to move from a passive role to the more active, engaged role that is required for fairness work. Additionally, we urge developers of \fautoml{} to present their system as what it is: an important tool amongst many others in the toolbox of a fairness-aware ML practitioner that can help them get the most out of the many possible methods for the modelling stage. In this spirit, developers should see \automl{} as a tool that primarily \textit{supports} practitioners when they engage in data work, rather than a tool that \textit{replaces} them \shortcite{wang-hci19a}.
Beyond communication, recognizing the narrow scope of current \automl{} systems also opens up paths for future work. For example, we envision \automl{} systems that provide user prompts or documentation that encourages practitioners to consider alternative interventions. More radically, future work could expand the frame of \automl{} to include more nuanced modeling of the sociotechnical system of which the ML model is a part, such as potential feedback loops, user interactions, or causal models.

\section{Effect of the AutoML System Design on Fairness}
\label{sec:automl_config}

Several design decisions of a \fautoml{} system shape its resulting models and for which tasks it will perform well. To illuminate the implications of those decisions, we discuss choices made for different components of the system including the \emph{search space}, \emph{formulation of the optimization problem} and \emph{evaluation protocol}.

\subsection{Search Space}
By designing the search space of an \automl{} system, we make implicit decisions about what kind of algorithms and what kind of processing steps a system should consider. This allows \emph{codifying established best practices}, e.g. by relying on processing steps and model classes that demonstrably lead to improved results.

This includes aspects such as feature selection and transformation methods, fairness-aware machine learning techniques, ML algorithms, and subsequent stacking or ensembling. To provide an example, the space of possible processing steps could be constrained to pre-processing steps and ML algorithms that yield easily interpretable ML pipelines, lending themselves to a better analysis by domain experts. While desirable, this may exclude pipelines that better reflect the real-world complexities captured in a dataset.
Sources of harm in this stage include \emph{aggregation bias}~\shortcite{suresh-eaamo21a}, which occurs when a learning algorithm is unable to capture the data distributions of distinct groups. Similarly, without considering fairness, extensive regularization may result in a model that only captures the data distribution of a majority group, but fails for minority groups.

Decisions made at this stage can also shape the capabilities of the resulting system that are not captured by fairness metrics. To provide an example, including fairness-aware machine learning techniques that return randomized predictions ~\shortcite<e.g.,>{hardt-nips16a,agarwal-icml18a} could lead to stochastic predictions in the final model. While this may lead to high performance on fairness metrics in expectation, randomization is an undesirable outcome in many real-world applications~\shortcite{weerts-arxiv22a}. 

\paragraph{Implications}
Components included in an AutoML system shape the quality of the resulting models and the system's applicability.
Focusing an \automl{} system's search space on inherently interpretable models might simplify and improve the reliability of model audits, but this might lead to models that are not sufficiently complex resulting in \emph{aggregation bias}. Similarly, allowing the user to include or exclude different components can lead to desirable or undesirable properties in the resulting models which should be reflected in the system's API design. For example, a \fautoml{} system could provide users flexibility in deciding whether to use techniques that lead to randomized predictions. An important research direction considers uncovering the advantages, disadvantages, and implicit normative assumptions of particular (fairness-aware) ML algorithms across multiple dimensions, including specific understandings of fairness and interpretability.

\subsection{Optimization Problem Formulation}
While previous work has used both constrained and multi-objective optimization to take fairness considerations into account, little attention has been given to the advantages and disadvantages that come with the two different paradigms, neither in the context of \fautoml{} nor in the general context of optimization.\footnote{We note that it is also possible to combine both paradigms into constrained multi-objective optimization, however, we are not aware of existing work that compares this approach to the constrained optimization and multi-objective optimization paradigms.}

Imposing a fairness constraint simplifies subsequent model selection procedures compared to the multi-objective scenario, as it allows for selecting the `best' model satisfying a previously specified constraint. This strongly simplifies the optimization problem~\shortcite{perrone-aies21a} and communication of results. On the other hand, by imposing constraints, we implicitly rephrase the goal of model building to \emph{find the best model that is still ethically/legally allowed}, ignoring the possibility to employ models that strike more favorable trade-offs. Additionally, how such constraints should be set is a non-trivial question -- particularly in advance. Previous work has often cited the \emph{four-fifths} rule~\shortcite<e.g.,>{feldman-kdd15a} as an example of such a constraint, but this rule only applies to a very narrow domain of US labor law and translating such legal requirements into fairness metrics requires multiple abstractions that likely invalidate resulting measurements~\shortcite{chen-arxiv22a}.\footnote{More generally, we argue that fairness metrics and constraints loosely derived from legal texts should not be equated with legal fairness principles.} Similarly, EU anti-discrimination law is designed to be context-sensitive and reliant on interpretation~\shortcite{wachter-clsr21a}, making it challenging to set any hard constraints in advance.

\begin{figure}[ht]
  \centering
  \begin{subfigure}{.5\textwidth}
  \centering
  \includegraphics[width=\textwidth]{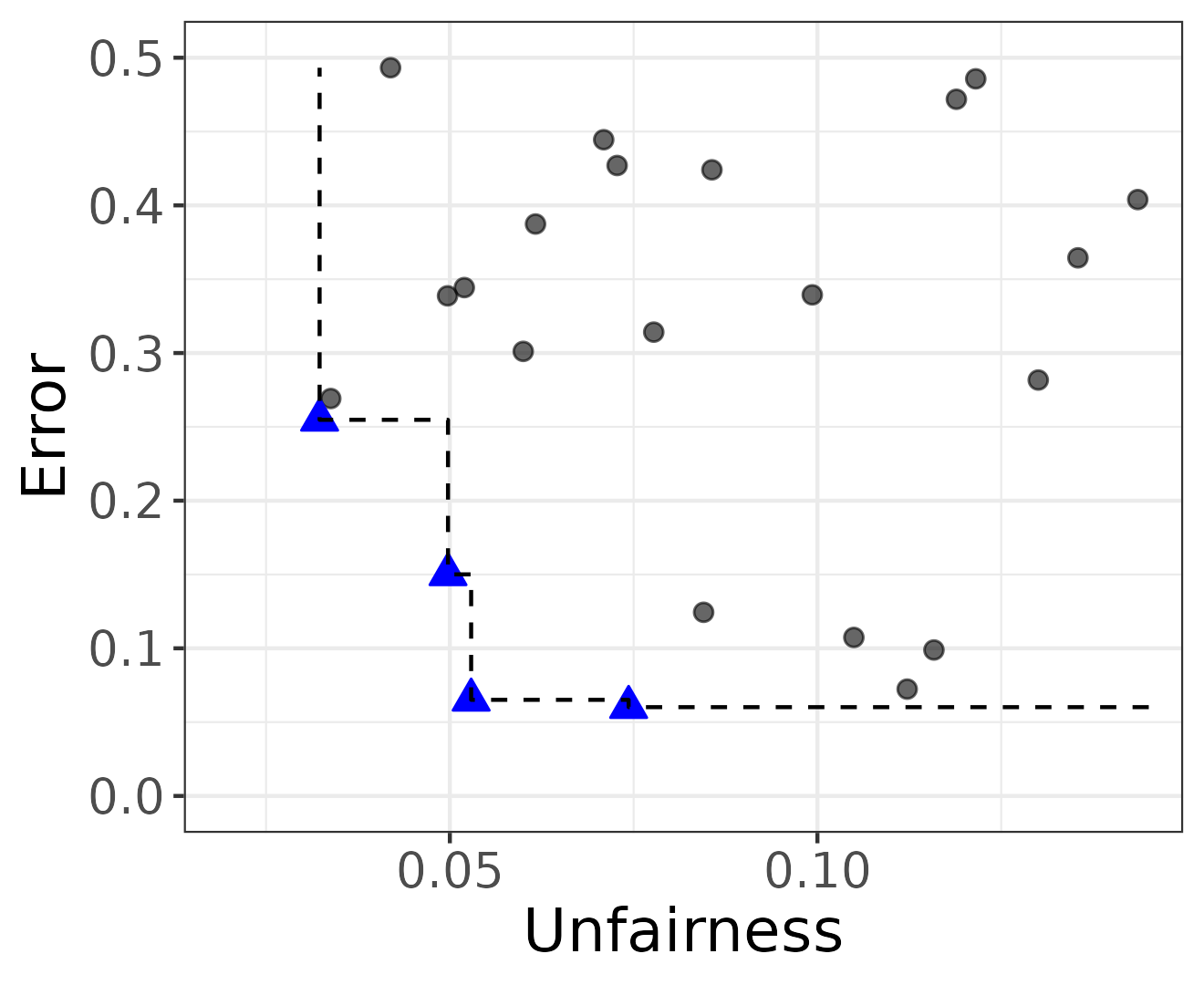}
  \caption{Multi-objective perspective}
  \label{fig:mo_s1}
\end{subfigure}%
\begin{subfigure}{.5\textwidth}
  \centering
  \includegraphics[width=\textwidth]{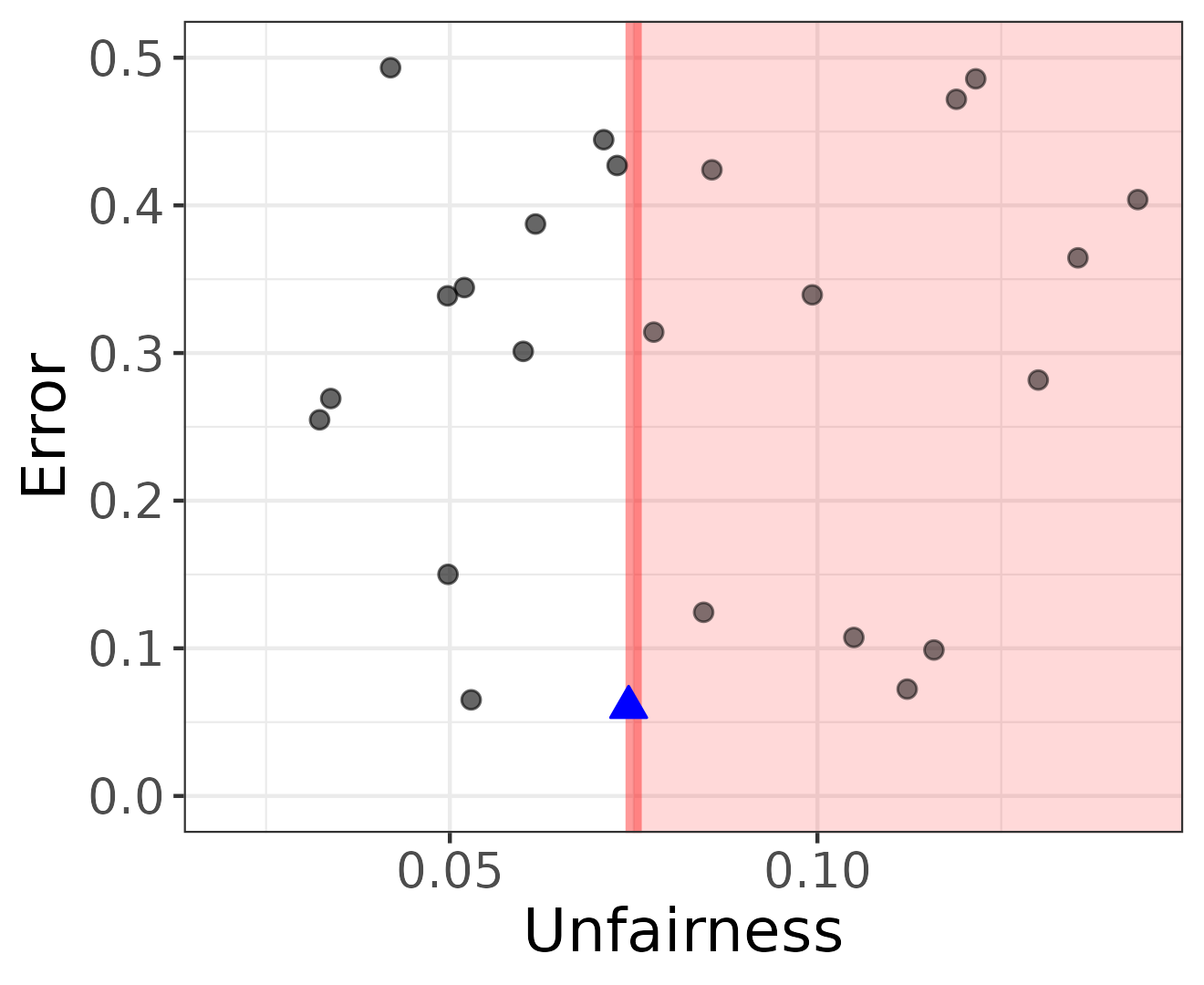}
  \caption{Constrained perspective}
  \label{fig:mo_s2}
\end{subfigure}
    \caption{Multi-objective perspective (left) and constrained perspective (right) showing exemplary unfairness and error values of found ML models (black dots). We depict optimal solutions returned by the minimization procedure as blue triangles, the approximation of the Pareto-front with a dashed line and the constraint (unfairness $>$ 0.075) as a red area.}
    \label{fig:mo_vs_cstrt}
\end{figure}

We visualize the multi-objective and constrained optimization perspective \wrt{} minimization in \Cref{fig:mo_vs_cstrt}. A multi-objective method (see \Cref{fig:mo_s1}) returns all solutions that are located on the Pareto-front (dashed line), i.e., solutions where no other more favorable trade-off is available. A constrained optimization method (see \Cref{fig:mo_s2}) instead disregards all solutions that violate a chosen constraint (here unfairness $<$ 0.075) depicted in red and returns the point with the lowest error from all solutions satisfying the constraint.

\paragraph{Implications}
Employing a multi-objective perspective allows for a better understanding of all available trade-offs, the effect of models and model hyperparameters, and which decisions impact fairness. This is particularly important since the fairness-aware (Auto)ML workflow is iterative in nature, due to the complexities of quantifying fairness and soliciting input from relevant stakeholders discussed in Section~\ref{sec:selecting_fairness}. Knowledge of the entire set of Pareto-optimal models is much more valuable in this iterative process than the single solution that a constrained optimization solution would yield. Furthermore, should constraints change in the future, results from multi-objective optimization might allow for selecting a different Pareto-optimal model while the constrained optimization approach might require full retraining of the entire \automl{} system. Further research into combining multi-objective and constrained optimization could lead to improved approximations of relevant segments of the Pareto-front without wasting compute budget on ethically questionable solutions that yield high performance.

\subsection{Evaluation Protocol}
Another set of problems arises from inadequate evaluation protocols. 
Most technical metrics applied during model selection only yield noisy estimates of the true generalization error due to the limited amount of data available and stochasticity of algorithms, soft- and hardware~\shortcite{bouthillier-mlsys21a}. The typical evaluation protocol uses a \emph{train-valid-test} split~\shortcite{raschka-arxiv18a}, where \emph{train} is used for training a model, \emph{valid} is used for model selection, i.e. selecting a final model from the \automl{} system, and \emph{test} is held out for a final evaluation, i.e. estimating the performance on unseen data or comparing \automl{} systems.\footnote{We note that instead of a single \emph{train-valid} split, one can also use other protocols, e.g. \emph{cross-validation}.} 

While the development of evaluation protocols for ML is challenging in general, the challenge is exacerbated when we consider group fairness metrics that require the estimation of group statistics (e.g., selection rate). Sensitive groups are often small, resulting in noisy estimates - especially when multiple sensitive characteristics are considered. Additionally, group fairness metrics summarize the comparison of those group statistics across groups (e.g., by taking the maximum difference), resulting in statistically biased upwards metrics that could possibly exaggerate disparities~\shortcite{lum-facct22}. 

Robust estimation is further complicated as evaluation often concerns multiple metrics (e.g., both a fairness constraint and a predictive performance objective). Estimating the performance of a given \fautoml{} system requires an evaluation protocol that ensures that the reported quantities (e.g. predictive performance or fairness metrics) are robustly estimated~\shortcite{agrawal-arxiv20a}.
The typical \emph{train-valid-test} protocol does not transfer to multiple objectives or additional constraints, where the performance of solutions is typically assessed as the hypervolume covered by non-dominated solutions -- the Pareto-optimal set~\shortcite{zitzler-tec03a}. Since metrics are noisy, candidate solutions chosen by the \automl{} system on \emph{validation} data might no longer satisfy constraints or be Pareto-optimal on the test set or when deployed in practice~\shortcite{feurer-ida23a}.

\paragraph{Implications}\label{sec:evalprotocol}
An ideal test set for fairness assessments is (1) free of measurement bias (see Section~\ref{sec:data}) and errors, and (2) contains sufficient data for each subgroup to accurately estimate group statistics. When subpopulations are very small, it may be infeasible to collect sufficient data through random sampling. In such cases, a weighted sampling approach in which small subgroups are oversampled may be more appropriate. However, care should be taken to not overburden already marginalized groups and to ensure that any further interpretation of the results takes note of sampling bias. If additional data collection is infeasible, stratification along the sensitive feature should be used to ensure that the test set approximately preserves the representation of each sensitive group in the train and test set - an approach commonly used for imbalanced classification problems. Future work should focus on developing evaluation protocols that explicitly take into account the uncertainty of estimates of fairness metrics. Furthermore, the assessment of multiple objectives or constraints requires the use of more robust evaluation protocols that are adapted to this setting \shortcite<e.g.,>{feurer-ida23a}.

\subsection{Benchmark-driven Development}
\label{sec:benchmark_dev}
Benchmarks for studying, developing and comparing algorithms have been widely used to track the state-of-the-art in subfields of ML research.
Benchmarks define typical tasks that should be solved by ML systems, providing a definition of each task's relevant properties, such as included data or evaluation metric(s).
In its simplest form, a benchmark task for \automl{} is characterized by a dataset (with a predefined set of features and target variable) and a predictive performance metric (including an evaluation protocol).
The design of \automl{} systems is often largely benchmark driven -- systems are developed to compete on standardized test suites~\shortcite{gijsbers-arxiv22a} or in AutoML competitions~\shortcite{guyon-automlbook19a}.
This has the benefit that new system components are immediately tested for their empirical performance and only included if they provide substantial benefits. 
Furthermore, this allows for thoroughly studying algorithms, objective comparisons, and visibility and reproducibility of research progress.
Translating benchmark results into real-world improvements requires the availability of and access to benchmarks that adequately reflect the contextual complexity and technical challenges of relevant real-world tasks~\shortcite{raji-neuripsdata21a}.
We identify several reasons why this assumption may not be applicable in the context of fairness research.

\paragraph{Existing benchmarks are decontextualized}
Several benchmarks have been critiqued for a lack of similarity to real-world applications~\shortcite{langley-ieee96a,saitta-ml98a,wagstaff-arxiv12a}. The construction of fairness benchmarks often seems to be guided by availability, rather than by a careful abstraction of a real-world problem. Especially in the context of fairness, these characteristics lead to issues around the validity of the collected data \shortcite{ding-neurips21a,bao-neuripsdbt21,gromping-techrep19a} as well as a disconnect from real-world applications \shortcite{bao-neuripsdbt21}. As a result, progress on fairness benchmarks is unlikely to reflect progress on the real-world outcomes that motivate fairness research.
For example, \shortciteA{bao-neuripsdbt21} describe how the use of ProPublica's COMPAS dataset~\shortcite{angwin-machinebias16a} as a benchmark is misleading, as performing well on this task cannot be tied to real-world impact in the field of criminal justice. Using benchmarks to solely chase state-of-the-art performance, researchers risk losing sight of the context of the data - which is crucial to advance social outcomes such as fairness.

\paragraph{Small set of benchmarks}
Many subfields in ML research focus on empirical performance improvements on a very small set of benchmark tasks, raising concerns regarding overfitting~\shortcite{recht-icml19a} and a lack of contribution to scientific understanding~\shortcite{hooker-jh95a,dehghani-arxiv21a}. Fairness-aware ML methods are no exception and have typically been evaluated only across a small set of benchmark tasks (cf. \Cref{tab:related_work_datasets} in the Supplementary material), strongly limiting the conclusions that can be drawn from such experiments on a meta-level.

\paragraph{Existing benchmark tasks are limited to the modeling stage}
The high emphasis on mitigation algorithms in algorithmic fairness research has resulted in a set of benchmark datasets (cf. \Cref{tab:related_work_datasets} in the Supplementary material) that only reflect a very small aspect of typical fairness-aware ML workflows since AutoML typically focuses on the modeling stage (cf. \Cref{fig:pipeline}). This can result in a blind spot for specific capabilities, such as evaluating long-term impact or handling intersectional sensitive groups, that are not tested in a typical fairness benchmark, which can trickle down to the users of the \automl{} system.

\paragraph{Implications}
If \fautoml{} system development is guided by benchmarks, the datasets making up such benchmarks should be \emph{contextualized}, \emph{reflect real-world challenges}, and \emph{sufficiently large}. We urge \automl{} and fairness researchers to jointly develop contextualized benchmark tasks that reflect the challenges that practitioners face in practice when working on fairness-related tasks. Defining such benchmark tasks requires a realistic application scenario along with problem constraints as well as relevant fairness and performance metric(s) tailored to real-world outcomes.
In many domains, data collection will require a great deal of effort that may take years, including (but not limited to) collaborating with domain experts, intense stakeholder management, and resolving incomplete and complex database systems.
Another important reason for the lack of datasets is concerns around data privacy, since fairness-related ML tasks typically comprise sensitive data. Synthetically generated data has been successfully used in other fields and could also provide a useful source here. This would enable researchers to thoroughly study the performance of algorithms \wrt{} to specific problem characteristics. However, even ``realistic'' synthetic datasets may not accurately reflect the real-world context compared to their ``real-world'' counterparts and care should be taken in the interpretation of the results. Additionally, data privacy issues could be tackled on an organizational level, but given the diversity of data protection regulations across jurisdictions, specific recommendations are beyond the scope of this work.
If constructed well, we believe that benchmarks could inspire the development of \fautoml{} systems with more diverse capabilities and highlight complexities and problems that occur in practice.

While quantitative benchmarks have widely been used to demonstrate a method's superiority, we argue that especially in the context of \fautoml{}, we need to move beyond numerical performance.
In practice, users might often prefer simpler solutions over complex methods as they simplify debugging and retraining. Moreover, performance gains from more complex models often do not translate to benefits during deployment~\shortcite{shankar-arxiv22a}. In particular, we believe future work should shift in focus from ``horse race'' analyses to a better understanding of when and why a particular system works and -- perhaps more importantly -- when it does not. Rather than pursuing incremental improvements in numerical performance, we believe that evaluation approaches designed to diagnose potential failure modes~\shortcite{raji-neuripsdata21a} and controlled experiments motivated by specific hypotheses~\shortcite{hooker-jh95a} will be crucial for advancing fairness research.

Until better benchmark tasks become available, we would like to highlight that we cannot endorse using existing benchmark tasks to claim progress in fairness research. If researchers use existing benchmark tasks to assess the strengths and weaknesses of multi-objective or constrained optimization algorithms, we recommend (1)~a clear statement that multi-objective / constrained optimization work does not constitute fairness research. We also recommend (2)~emphasizing work on more sensible benchmarks and (3)~relying on very targeted problem scenarios within a clearly defined application context, clearly stating involved biases and resulting harms.

\section{Opportunities for Fairness-aware \automl{}}
\label{sec:opportunities}
Provided that the use of (fairness-aware) ML is justifiable, ML pipelines should be constructed in the best possible manner. Existing work points towards several opportunities for \fautoml{} to contribute to this goal.

\paragraph{\Fautoml{} allows integrating best practices in the modeling stage.}
Practitioners often lack knowledge on how to integrate fairness toolkits into their workflow~\shortcite{holstein-chi19a,deng-facct22a} and may face difficulties in proper evaluation of fairness metrics~\shortcite{agrawal-arxiv20a}.
\automl{} systems can implement relevant search spaces, mitigation techniques, and evaluation protocols into a configurable pipeline that is optimized automatically, relieving the user from staying up-to-date with the technical literature.
Moreover, \automl{} systems can codify best practices that ensure that model selection is done correctly, preventing mistakes arising from, for example, wrongly coded evaluation protocols or undetected train-test leakage.

\paragraph{\Fautoml{} may outperform traditional fairness-aware ML techniques.}
Previous work has shown that (1)~even an arguably simple constrained Bayesian optimization of a standard ML algorithm is en-par with solutions discovered by fairness-aware ML algorithms~\shortcite{perrone-aies21a}, with the advantage of being much more flexible, and that (2)~the joint multi-objective optimization of deep neural architectures and their hyperparameters can also outperform other bias mitigation techniques in the context of face recognition~\shortcite{dooley-neurips23a}.
Importantly, standard fairness-aware ML algorithms, such as a fairness-aware support vector machine~\shortcite{donini-neurips18a} require hyperparameter tuning, too, which is an additional argument for employing \fautoml{} techniques.
We are not aware of a thorough comparison of manual tuning with a constrained or multi-objective hyperparameter optimization algorithm, but by extrapolating results from the single-objective setting~\shortcite{bergstra-jmlr12a,snoek-nips12a,chen-arxiv18a,henderson-aaai18a,zhang-aistats21a} 
into a constrained or multi-objective setting, 
we believe that \fautoml{} is perfectly posed to support practitioners to reliably and efficiently tune hyperparameters.

\paragraph{\Fautoml{} may fit more easily into existing workflows.}
By building upon standard HPO rather than hand-crafted ML algorithms that were altered to take fairness into account, AutoML solutions are typically easier to fit into existing workflows~\shortcite{cruz-icdm21a} and can be used together with different off-the-shelf machine learning models.
For example, \shortciteA{perrone-aies21a} demonstrate that their method works for random forests~\shortcite{breimann-mlj01a}, XGBoost~\shortcite{chen-kdd16a}, neural networks, and linear models. Similarly, \shortciteA{cruz-icdm21a} have jointly optimized random forests, decision trees, LightGBM~\shortcite{ke-neurips17a}, logistic regression, neural networks, and decision trees wrapped with the exponentiated gradient method~\shortcite{agarwal-icml18a} -- also showing that each of the models leads to a different trade-off between the performance and the fairness metric.
In addition (and in contrast to most fairness-aware ML strategies), \fautoml{} systems are typically agnostic to the fairness metric at hand~\shortcite{perrone-aies21a,chakraborty-ase19a,cruz-icdm21a,wu-arxiv21a} and can be used for arbitrary models while not relying on a specific model class being adapted~\shortcite{perrone-aies21a}.
As such, \fautoml{} has the potential to address the portability trap described by~\citeA{selbst-fat19a}: while individual models and even individual bias mitigation techniques do not generalize from one usage context to another, the AutoML process of \emph{searching} for the Pareto front of optimal tradeoffs of user-defined objectives in a particular application generalizes across usage contexts.
Moreover, algorithmic solutions are sometimes deployed without considering fairness issues at all, due to a lack of awareness, knowledge, or expertise~\shortcite{lum-s16a,buolamwini-facct18a}. While technical interventions should not be regarded as the sole tool for addressing unfairness, enriching \automl{} can lower the barrier for practitioners and domain experts to incorporate fairness considerations. Relieving the user from some of the technical complexities of ML pipeline design can allow them to spend more time on aspects where a human in the loop is essential.

\paragraph{\Fautoml{} can be a useful tool to improve understanding of metrics, data, \& models}
Machine learning can be a useful tool not just during model development, but also during data exploration. In particular, ML can be used to investigate relationships in the data that lead to potential fairness-related harm~\shortcite{wachter-vlr2020a}. \automl{} can be a useful tool to quickly build a variety of ML models with high predictive performance. By evaluating models \wrt{} a wide variety of (fairness) metrics, practitioners can gain insight into potential biases contained in the data and labels. While current \automl{} systems typically do not facilitate this type of analysis, we envision future systems that allow a practitioner to investigate which features a model typically relies on or whether an (unconstrained) ML pipeline leads to performance disparities. Such knowledge can then be used to, e.g. inform policy decisions that address fairness-related harms at their core.
By using a multi-objective \automl{} system for exploration, a user can also learn about the trade-offs between different objectives. In contrast to comparing solutions obtained by manual inspection, a multi-objective \automl{} system directly explores the Pareto front of optimal trade-offs.

\paragraph{\Fautoml{} solutions lend themselves to further inspection of the learning process.}
Such inspections have traditionally been used to better understand the relationship between the characteristics of an ML task, the parameters of learners, and predictive performance \shortcite{jones-jgo98a,HutHooLey13c,hutter-icml14a,fawcett-heu16a,golovin-kdd17a,biedenkapp-aaai17a,biedenkapp-lion18a,rijn-kdd18a,probst-jmlr19a,weerts-arxiv20a,moosbauer-neurips21a}. Similar analyses can be performed in fairness-aware ML. For example, \shortciteA{perrone-aies21a} used hyperparameter importance analysis to get further insights into the learners, finding that regularization parameters are particularly important to meet fairness constraints. 
These insights are readily available as output of an \automl{} system, having generated meta-data during the interaction of pipeline optimization and the data at hand.\\

\section{Conclusions}\label{sec:conclusion}
\Fautoml{} systems have the potential to overcome several challenges ML practitioners face in fairness work. By simplifying and accelerating the integration of fairness considerations into ML solutions, these systems can allow practitioners to allocate more time to tasks where a human-in-the loop is essential. Moreover, \fautoml{} systems can be a useful tool to rapidly improve understanding of the input data and machine learning process. However, the effectiveness of \fautoml{} systems in mitigating real-world harm depends on various factors, including not only the design of the system, but also the way in which practitioners utilize it. Furthermore, while issues related to fairness can emerge in every stage of the ML workflow, \automl{} is currently primarily involved with the modeling stage, limiting its role in fairness-aware ML workflows. This raises the question: how can we develop \fautoml{} systems in a way that acknowledges the complexity of algorithmic fairness throughout the entire workflow? In this concluding section, we propose several guidelines for research and development of \fautoml{} and set out directions for future work.

\paragraph{Guidelines for Fairness-aware AutoML}
\label{sec:guideslines}
It is important to be aware of the consequences of making technological solutions available. To assist researchers and developers interested in \fautoml{}, we have formulated the following guidelines.
\begin{itemize}
    \item \textbf{Clearly state assumptions and limitations.} When talking about \fautoml{} and solutions generated by such systems, prominently state assumptions and limitations. It must be clear for what use cases the system is suitable, how it was tested, and in which scenarios it should not be used. In particular, clearly state that satisfying fairness metrics cannot guarantee a fair system~\shortcite{corbett-arxiv18a} and may even lead to adverse effects~\shortcite{cooper-aies21a}.
    \item \textbf{Support users in identifying sources of fairness-related harm.} 
    Sources of fairness-related harm can often be mitigated more effectively through interventions beyond the modeling stage of an ML workflow, such as better data collection, an alternative problem formulation, and non-technical interventions related to the use of the system. As such, it is important to help users identify potential issues. One direction for this could be informing users of potential problems via fairness checklists~\shortcite{madaio-chi20a} or inviting explicit reflection on data and model lineage, e.g., by pre-populating or simplifying the creation of \emph{Data sheets}~\shortcite{gebru-acm21a} and \emph{Model cards}~\shortcite{mitchell-facct19a}.
    \item \textbf{Incorporate principles of seamful design.}
    Automation involves the risk that users overly trust solutions without employing the required amount of scrutiny of data and resulting models. This might result in sources of bias not accounted for during model development, evaluation, and deployment and, thus, could ultimately lead to adverse effects for sensitive groups. Systems should therefore take great care \wrt{} how solutions are presented and which conclusions can be drawn from reported metrics. To counteract automation bias, we encourage system developers to consider existing practices in human-computer interaction, such as adding intentional friction through seamful design~\shortcite{kaur-facct22a}, which encourages users to reflect on explicit and implicit design choices.
    \item \textbf{Support users in statistically sound fairness evaluation.}
    Assessing the fairness of a model requires that quantities are estimated based on a sufficiently large and representative sample of the population under consideration while ensuring that the data can support such conclusions.
    If not caught, errors in the evaluation could lead to blind spots for subgroups not sufficiently represented in the data or premature conclusions that are unsupported by the available data. \automl{} can and should codify best practices to prevent user errors during evaluation. In particular, we recommend warning users if the results of a fairness evaluation are not supported by sufficient data.
    \item \textbf{Account for inherent limitations of fairness metrics.}
    It is challenging to capture nuanced understandings of fairness in quantitative fairness metrics. As a result, optimization for a particular fairness constraint or objective can have undesirable side effects. To avoid these issues, \fautoml{} should support users in comprehensive evaluations beyond simple metrics. Additionally, we discourage solutions that enforce bias-transforming metrics such as demographic parity without careful modeling of the social context and data collection processes that motivate these metrics.
    \item \textbf{Ensure well-substantiated system design.}
    The search space is an important design choice of a \fautoml{} system. It defines the applicability and quality of models an \automl{} system outputs, as well as the trade-offs between desiderata that will be explored during learning. We recommend carefully documenting the design decisions made during the development process and explaining how they incorporate user requirements.
    \item \textbf{Evaluate the system against contextualized benchmarks.}
    If new \fautoml{} systems are designed, it is important that those systems not only solve an oversimplified problem but actually assist the user in achieving fairer outcomes. Existing benchmarks are often oversimplified and do not reflect the real-world requirements of a \fautoml{} system. Hence, novel contextualized benchmarks should be created to analyze to what extent \fautoml{} systems meet these requirements.
    \item \textbf{Support users in performing quick iterations.} The fairness-aware (Auto)ML workflow is necessarily iterative. We thus encourage the development of fast and interactive \fautoml{} systems that allow for rapid iterations.
\end{itemize}

\paragraph{Opportunities}
Having discussed perspective of users and developers of \fautoml{} systems in Sections~\ref{sec:user} and ~\ref{sec:automl_config}, as well as the opportunities for \fautoml{} systems in Section~\ref{sec:opportunities}, we now briefly summarize the potential benefit of \fautoml{} for each stakeholder to open up a dialogue for progressing towards \fautoml{}:
\begin{itemize}
    \item \textbf{\automl{} researcher.} \automl{} systems can have a positive real-world impact if they support users in building fairer models. Directly applicable research directions are improving evaluation and constrained multi-objective optimization. Moreover, we suggest extending the limited knowledge on how users interact with \automl{} systems~\shortcite{xin-chi21a} and developing interactive systems that support them best in doing so~\shortcite{wang-hci19a,crisan-chi21a}.
    \item \textbf{Fairness researcher.} Fairness-aware \automl{} stands and falls with the quality of available fairness metrics~\shortcite{ruggieri-aaai23a} and available testbeds to develop \fautoml{} systems. Therefore, we expect further research in contextual benchmark problems in combination with less abstract and more tangible fairness criteria will be a fruitful direction for future research. 
    \item \textbf{ML practitioner.} We urge users of ML to familiarize themselves with the dangers of applying ML in sensitive applications, for example by following the references provided in this paper. If designing an ML intervention is the best solution to a problem, we propose relying on principled techniques for building ML models and suggest \fautoml{} techniques. 
\end{itemize}

\paragraph{Future Work}
We posit that \fautoml{} systems can play an important role in typical fairness-aware ML workflows by tackling model selection, hyperparameter optimization, and model evaluation and identify several promising directions for future work towards this end.
An important research topic is the design of evaluation protocols that can properly handle multiple objectives or constraints, as well as noisy metrics in a statistically sound way. Furthermore, while most current work on fairness-aware ML approaches the problem from a constrained optimization perspective, much less work has explored the promising multi-objective optimization scenario.
Moreover, incorporating aspects of the \textit{evaluation} stage into the AutoML loop, by e.g., modeling latent preferences expressed during evaluation or making it easier to incorporate other real-world objectives into the optimization loop, is a promising avenue for further research.
We also believe that \automl{} and hyperparameter analysis can be a useful tool for fairness researchers as well as \automl{} researchers to improve understanding of the (fairness-aware) learning process and inform guidance on the suitability of particular approaches.
Additionally, we hope to encourage the joint development of more realistic benchmarks.
Furthermore, the present paper focuses primarily on settings with structured data, and we are aware of only one publication in the realm of \fautoml{} that tackles image data~\shortcite{dooley-neurips23a}. While many of the discussed challenges are also relevant to settings with unstructured data, we expect several unique challenges related to bias and unfairness of models trained on unstructured data such as images and text, for example, missing meta-data on group membership. We believe that exploring these challenges and opportunities is an interesting direction for future work.
Similarly, generative AI shares many of the challenges we outlined, but will require additional considerations that warrant further research.
Finally, we want to argue in favor of \automl{} systems that are more interactive in nature: instead of monolithic one-solution-fits-all systems, future work should go into \emph{assistive} systems, that, e.g., point out possible problems in the data, or guide the user through fairness audits \shortcite{landers-ap22a,madaio-chi20a}. We believe that an interdisciplinary approach bridging \automl{} and human-computer interaction research will be crucial for designing effective user interaction interfaces and counteracting potential automation bias. \\

\noindent Revisiting the original question that motivated this work, we conclude that \emph{fairness cannot be automated}. Instead, an iterative, context-sensitive process guided by human domain experts is essential to arrive at fair outcomes. 
Nevertheless, in cases where technical interventions are appropriate, \emph{\fautoml{} systems can lower the barrier to incorporating fairness considerations} in the ML workflow and support users through the \emph{integration of best practices and state-of-the-art approaches}, without the need to follow the latest technical literature. 
However, we emphasize that basic data science skills remain crucial for correctly framing the problem and continuously monitoring performance.
In summary, even with a humble attitude towards the role of \automl{}, we believe \fautoml{} to be an important research direction that is likely to have a substantive impact.

\subsection*{Acknowledgement}
Hilde Weerts and Florian Pfisterer contributed equally to this work.
Matthias Feurer, Katharina Eggensperger, Noor Awad and Frank Hutter acknowledge the Robert Bosch GmbH for financial support.
Katharina Eggensperger also acknowledges funding by the German Research Foundation under Germany's Excellence Strategy - ECX number 2064/1 - Project number 390727645.
Edward Bergman, Joaquin Vanschoren and Frank Hutter acknowledge TAILOR, a project funded by EU Horizon 2020 research and innovation programme under GA No 952215. 
Edward Bergman, Noor Awad and Frank Hutter acknowledge funding by the European Union (via ERC Consolidator Grant DeepLearning 2.0, grant no.~101045765). Views and opinions expressed are however those of the author(s) only and do not necessarily reflect those of the European Union or the European Research Council. Neither the European Union nor the granting authority can be held responsible for them. \begin{center}\includegraphics[width=0.3\textwidth]{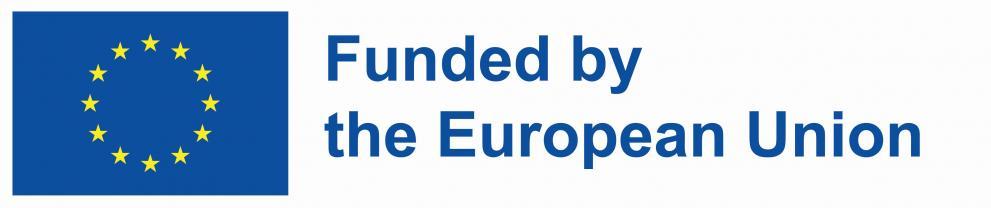}\end{center}

\appendix
\section{Datasets Used in \Fautoml{} Research} 
We collect datasets used for empirical comparisons in the \fautoml{} works we cite in order to highlight the limited scope and quantity of datasets used during benchmarks. \fautoml{} systems are typically evaluated on only $1$ to $4$ datasets, typically employing either a fairness metric and a performance metric (A-E) or a combined metric that evaluates performance given a fairness constraint is satisfied (F-I). While additional datasets have been collected and investigated in the context of fairness benchmarking \shortcite{quy-wires22a,fabris-dmkd22}, they have not been used to evaluate \fautoml{} systems. In addition, benchmarks based on such datasets might suffer from the same problems, such as limited scope and lack of representativity discussed in \Cref{sec:benchmark_dev}.
\label{sec:prevdata}
\begin{table}[htbp]
    \centering
    \small
    \caption{Fairness-related datasets used in prior research on \fautoml{}.}
    \label{tab:related_work_datasets}
    \begin{tabular}{
    l@{\hskip 2mm}
    c@{\hskip 2mm}
    c@{\hskip 2mm}
    c@{\hskip 2mm}
    c@{\hskip 2mm}
    c@{\hskip 2mm}
    c@{\hskip 2mm}
    c@{\hskip 2mm}
    c}
    \toprule
    \multirow{4}{*}{Paper} & 
    \multirow{4}{*}{Adult} & 
    \multirow{4}{1.3cm}{German Credit} &
    \multirow{4}{*}{Compas} & 
    \multirow{4}{1.3cm}{Donors Choice} & 
    \multirow{4}{1.3cm}{AOF (private)} & 
    \multirow{4}{1.3cm}{Default Risk (Kaggle)} &
    \multirow{4}{*}{Bank} & 
    \multirow{4}{*}{MEPS} \\
    & & & & & & & \\
    & & & & & & & \\
    & & & & & & & \\
    \midrule
    \textbf{A} \shortciteA{pfisterer-arxiv19a} & \checkmark & & & & & & & \\
    \textbf{B} \shortciteA{schmucker-metalearn20a} & & \checkmark & \checkmark & & & & & \\
    \textbf{C} \shortciteA{schmucker-arxiv21a} & \checkmark & & & & & & & \\
    \textbf{D} \shortciteA{chakraborty-ase19a} & \checkmark & \checkmark & \checkmark & & & & & \\
    \textbf{E} \shortciteA{cruz-icdm21a} & \checkmark & & \checkmark & \checkmark & \checkmark & & & \\
    \textbf{F} \shortciteA{liu-aaai20a} & & & & & & \checkmark & & \\
    \textbf{G} \shortciteA{perrone-aies21a} & \checkmark & \checkmark & \checkmark & & & & & \\
    \textbf{H} \shortciteA{wu-arxiv21a} & \checkmark & & \checkmark & & & & \checkmark & \\
    \textbf{I } \shortciteA{wu-arxiv21b} & \checkmark & & \checkmark & & & & \checkmark & \checkmark\\
    \bottomrule
\end{tabular}
\end{table}

\bibliography{strings,lib,local,proc}
\bibliographystyle{theapa}

\end{document}

%% file: macros.tex
\newcommand{\wrt}{w.r.t.}

\newcommand{\conf}[0]{\bm{\lambda}}

\newcommand{\automl}{AutoML}
\newcommand{\fautoml}{fairness-aware \automl{}}
\newcommand{\Fautoml}{Fairness-aware \automl{}}
\DeclareMathOperator*{\argmin}{arg\,min}


%% file: figures/workflownew.tex
\definecolor{dblue}{RGB}{55,126,184}
\definecolor{dorange}{RGB}{255,176,0}

\begin{figure}[ht]
    \centering
\tikzset{%
  >={Latex[width=2mm,length=2mm]},
    base/.style = {rectangle, draw=black,
                   minimum width=2.5cm, minimum height=1cm,
                   text centered, solid},
    process/.style = {base, rounded corners, minimum width=2.5cm, solid},
    choice/.style = {base,  minimum width=2.3cm, fill=dblue!45, minimum height=1.2cm}
}
\resizebox{\textwidth}{!}{
\begin{tikzpicture}[node distance=4cm, ->, line width=0.8, dotted, auto,
    every node/.style={fill=white, font=\sffamily}, align=center]

  \node (problem)   [process] {1. Problem \\ understanding};      
  \node (data)      [process, right of = problem, xshift=-0.15cm] {2. Data understanding \\ and preparation};
  \node (model)     [process, right of = data, xshift=0.95cm] {3. Modeling};
  \node (eval)      [process, right of = model, xshift=0.3cm] {4. Evaluation};
  \node (deploy)    [process, right of = eval, xshift=-0.7cm] {5. Deployment};
  \draw[<->]             (problem) -- (data);
  \draw[<->]             (data) -- (model);
  \draw[<->]             (model) -- (eval);
  \draw[<->]             (eval) -- (deploy);

  \node (config) [above of = model, xshift=0cm, yshift=-2.0cm] {
    \textbf{System Configuration} \\
    search space \& strategy, \\
    optimization formulation, \\
    evaluation protocol
    };
  \draw[thick,dashed] ($(config.north west)+(-0.2,0.1)$)  rectangle ($(config.south east)+(0.2,-0.1)$);

  \node (user) [circle, draw=black, solid, below of = model, yshift=1.75cm] {user};
  \draw[->, solid] (user) -| ($(user)+(-0.8,0)$) -- ($(model.south)+(-0.8,0)$);
  \node (input) [left of = user, xshift=2.4cm, yshift =0.6cm ] {
    data, \\
    metrics \\
  };
  \draw[<-, solid] (user) -| ($(user)+(0.8,0)$) -- ($(model.south)+(0.8,0)$);
  \node (output) [right of = user, xshift=-2.4cm, yshift=0.9cm ] {model};
  \node (interact) [below of = user, xshift=0cm, yshift=3.1cm] {\textbf{User Input \& Interaction}};
  \draw[thick,dashed] ($(interact.north west)+(-0.1,2.1)$) rectangle ($(interact.south east)+(0.1,-0.1)$);

  \node (scope) [above of = model, xshift=0cm, yshift=-0.3cm] {\textbf{AutoML Scope}};
  \draw[thick,dashed] ($(scope.north west)+(-1.2,0.1)$) rectangle ($(scope.south east)+(1.2,-7.05)$);
  
  \node (review)    [choice, below of = problem, yshift=-1cm] {Ethical Review \& \\ System (Re)design};
  \node (ident)     [choice, below of = data, yshift=-1cm] {Bias Identification \& \\ Mitigation};
  \node (mitig)     [choice, below of = model, yshift=-1cm] {Fairness-aware\\ Machine Learning};
  \node (assess)  [choice, below of = eval, yshift=-1cm, xshift=1.8cm] {Continuous \\ Fairness Assessment};
  \draw[solid,->]             (review) -- (problem);
  \draw[solid,->]             (ident) -- (data);
  \draw[solid,->]             (mitig) -- ($(model)-(0,3.7)$);
  \draw[solid,->]             (assess) -- (eval);
  \draw[solid,->]             (assess) -- (deploy);
  \node (stake) [choice, below of = model, xshift=0cm, yshift = -3cm] {Stakeholder Involvement};
  \draw[solid,->]             (stake) -| (review);
  \draw[solid,->]             (stake) -| (ident);
  \draw[solid,->]             (stake) -| (assess);
  \draw[solid,->]             (stake.north) -| (mitig);
  
  \draw[-]             (eval.north) -- +(0,4);
  \draw[-]             (deploy.north) -- +(0,4);
  \draw[->]             (deploy.north) +(0,4) -| (problem.north);
  \end{tikzpicture}
 }
    \caption{Example of an ML workflow adapted from the CRISP-DM~\protect\shortcite{shearer-jdw00a} process. 
     Developing ML systems is an iterative process (dotted arrows) that can require frequently revisiting decisions made in previous stages. Most existing fairness-aware AutoML methods and systems address fairness analogous to fairness-aware ML techniques: the problem is formulated as an optimization task under a fairness objective or constraint. However, many important design choices are made outside of the \textit{modeling} stage which is typically the part of the workflow that is tackled by AutoML systems. Taking fairness into account adds additional considerations (in blue) to every step of the ML workflow. }
    \label{fig:pipeline}
\end{figure}